\pdfoutput=1

\documentclass[11pt]{article}

\usepackage{acl}

\usepackage{times}
\usepackage{latexsym}
\usepackage{graphicx}
\usepackage{subcaption}
\usepackage{caption}

\usepackage[utf8]{inputenc}

\usepackage{booktabs}

\usepackage{bbding}
\usepackage{float}
\usepackage{pifont}
\newcommand{\xmark}{\ding{55}}%

\usepackage[T1]{fontenc}

\usepackage[utf8]{inputenc}

\usepackage{microtype}

\usepackage{enumitem}
\usepackage{multirow}
\usepackage{xcolor}
\usepackage{xspace}

\usepackage[capitalize,noabbrev]{cleveref}
\crefname{chapter}{Chapter}{Chapters}
\crefname{section}{\S}{\S\S}
\Crefname{section}{\S}{\S\S}
\crefname{table}{Table}{Tables}
\crefname{figure}{Figure}{Figures}
\crefname{algorithm}{Algorithm}{}
\crefname{equation}{Eq.}{}
\crefname{appendix}{Appendix}{}
\crefformat{section}{\S#2#1#3}

\newif\ifdraft
\drafttrue

\title{
LogogramNLP: Comparing Visual and Textual Representations \\
of Ancient Logographic Writing Systems for NLP
}
\author{
  Danlu Chen$^{1}$,
  Freda Shi$^{2}$,
  Aditi Agarwal$^{1}$,
  Jacobo Myerston$^{1}$,
    Taylor Berg-Kirkpatrick$^{1}$ \\
  UC San Diego$^{1}$, University of Waterloo$^{2}$ \\
  \texttt{danlu@ucsd.edu}
}

\begin{document}
\maketitle
\begin{abstract}

  Standard natural language processing (NLP) pipelines operate on symbolic representations of language, which typically consist of sequences of discrete tokens.
  However, creating an analogous representation for ancient logographic writing systems is an extremely labor-intensive process that requires expert knowledge.
  At present, a large portion of logographic data persists in a purely visual form due to the absence of transcription---this issue poses a bottleneck for researchers seeking to apply NLP toolkits to study ancient logographic languages: most of the relevant data are \textit{images of writing}.
  This paper investigates whether direct processing of visual representations of language offers a potential solution.
  We introduce \textbf{LogogramNLP}, the first benchmark enabling NLP analysis of ancient logographic languages, featuring both transcribed and visual datasets for four writing systems along with annotations for tasks like classification, translation, and parsing. 
  Our experiments compare systems that employ recent visual and text encoding strategies as backbones.
  The results demonstrate that visual representations outperform textual representations for some investigated tasks, suggesting that visual processing pipelines may unlock a large amount of cultural heritage data of logographic languages for NLP-based analyses. Data and code are available at \url{https://logogramNLP.github.io/}.
\end{abstract}

\section{Introduction}


The application of computational techniques to the study of ancient language artifacts has yielded exciting results that would have been difficult to uncover with manual analysis alone \citep{assael2022restoring}.
Unsurprisingly, one of the biggest challenges in this domain is data scarcity, which, in turn, means that transferring from pre-trained systems on well-resourced languages is paramount.
However, it is more challenging to adopt similar techniques for ancient logographic writing systems, in which individual symbols represent entire semantic units like morphemes or words.

\begin{figure}[t]
    \includegraphics[width=0.48\textwidth]{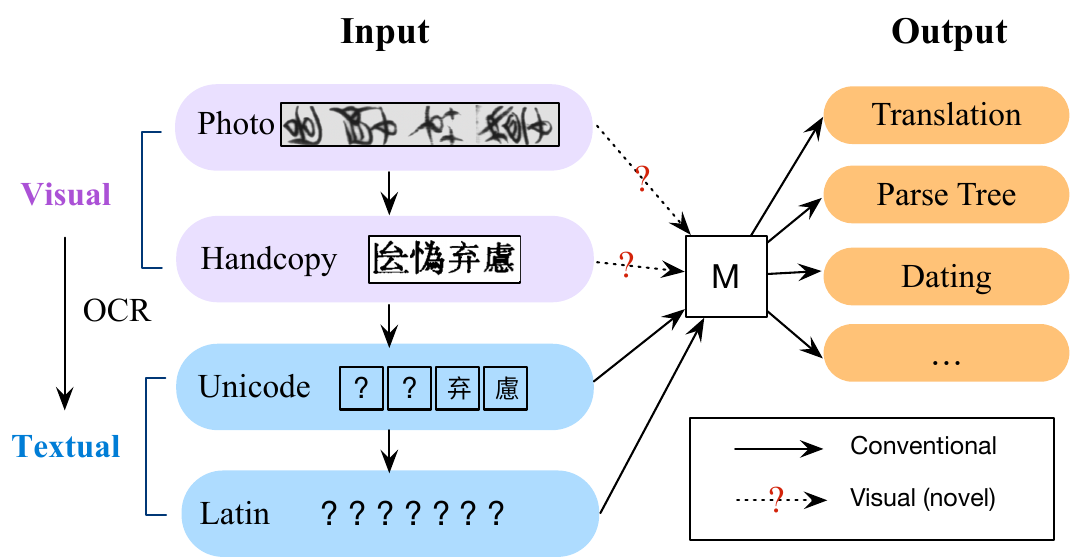}
    \caption{\label{fig:old_chinese}
        Illustration of the processing flow of Old Chinese (in Bamboo Script), an ancient logographic language, best viewed in color.
        M denotes the pre-trained model used in the pipeline.
        Vision-based models directly process visual representations (violet; dashed lines).
        Conventional NLP pipelines (blue; solid lines) first convert visual representations into symbolic text---either automatically, which is quite noisy, or manually, which is labor-intensive.
        However, as shown, some ancient logographic writing systems have symbol inventories that have not yet been fully mapped into Unicode.
        Even when Unicode codepoints exist, they are often mutually exclusive with the symbol inventories of high-resource languages, reducing the effectiveness of transferring from pre-trained models.
        Finally, \textit{latinization} (a potential solution for finding common ground with pre-training languages) loses information from the original input, is not fully standardized, and is difficult to automate.
    }
\end{figure}

The challenges associated with NLP for ancient logographic languages mainly come from two aspects.
First, for many ancient languages, most available data sources are in visual forms, consisting of untranscribed photographs or hand-drawn copies (i.e., \textit{lineart}).
Adopting the conventional NLP pipeline, which requires converting visual representations into symbolic text, is therefore not straightforward: automatic transcriptions are often noisy due to data scarcity, while manual transcriptions are labor-intensive and require domain expertise.
Some logographic writing systems, such as Old Chinese, even include symbol inventories that remain not fully mapped to Unicode (depicted in \cref{fig:old_chinese}).

Second, even when perfect Unicode transcriptions are available, their symbol inventories are often mutually exclusive with those of high-resource languages, which can substantially reduce the effectiveness of transfer from pre-trained multilingual encoders, such as mBERT \citep{devlin2018bert}.
One processing step that might be used to mitigate this issue is \textit{latinization} of the Unicode transcripts \citep{rust-etal-2021-good,muller2020unseen}.
However, it is challenging to Latinize logographic languages due to uncertain pronunciations \citep{sproat2021taxonomy} and the resulting inconsistent latinization schemes across artifacts from the same language and writing system.
Such a process is laborious---humanists may devote months or even years to determine the correct transliteration.
In contrast, once a correct transliteration is determined, translation into another language may only take minutes.

\begin{table*}[t]
    \resizebox{\linewidth}{!}{
        \begin{tabular}{@{}llccccccccc@{}}
            \toprule
            \multirow{2}{*}{\textbf{Writing system}}                   & \multirow{2}{*}{\textbf{Language}} & \multirow{2}{*}{\textbf{abbr.}} & \multicolumn{2}{c}{\textbf{Visual Feature}} & \multicolumn{2}{c}{\textbf{Textual Feature}} &                        & \multicolumn{3}{c}{\textbf{Task}}                                                                                        \\
            \cmidrule(lr){4-5}  \cmidrule(lr){6-7} \cmidrule(lr){9-11} &                                    &                                 & \textbf{Full Doc}                           & \textbf{Textline}                            & \textbf{Unicode}       & \textbf{latinization}             &  & \textbf{Translation}         & \textbf{UD Parsing} & \textbf{Attribute}           \\ \midrule
            \textbf{Linear A}                                          & Unknown                            & LNA                             & Y                                           & \underline{\textbf{Y}}                       &                        & \underline{\textbf{Y}}            &  &                              &                     & \underline{\textbf{Y}}       \\
            \textbf{Egyptian hieroglyph }                              & Ancient Egyptian                   & EGY                             &                                             & \underline{\textbf{Y}}                       &                        & \underline{\textbf{Y}}            &  & \underline{\textbf{Y}}       &                     & \underline{\textbf{Y}}       \\
            \textbf{Cuneiform}                                         & Akkadian \& Sumerian               & AKK                             & Y                                           &                                              & Y                      & Y                                 &  & Y                            & Y$^{*}$             & Y                            \\
            \textbf{Bamboo script}                                     & Ancient Chinese                    & ZHO                             &                                             & \textbf{\underline{Y}}                       & \underline{\textbf{Y}} & \underline{\textbf{Y}}            &  & \underline{\textbf{Y$^{*}$}} &                     & \underline{\textbf{Y$^{*}$}} \\ \bottomrule
        \end{tabular}
    }
    \caption{A summary of the task availability across four ancient languages with unique writing systems. The \underline{\textbf{underlined}} \underline{Y} indicates that the data has not previously been used in a machine learning setup, which demonstrates the novelty of our benchmark; and asterisks ($^{*}$) indicate that we conducted extra manual labeling.
    }
    \label{tb:tasksum}
\end{table*}

Fortunately, advances in visual encoding strategies for NLP tasks offer an alternative solution.
Recent studies have investigated NLP systems that model text in the pixel space \citep{rust-etal-2023-pixel,tschannen2023clippo,salesky-etal-2023-multilingual}, thereby opening new possibilities for the direct use of visual representations of ancient logographic writing systems.
These approaches, to date, have primarily been applied to digitally rendered texts.
They have not yet been extensively evaluated on handwritten texts, such as \textit{lineart}, i.e., neatly hand-copied versions of texts by scholars.

In this paper, we attempt to answer the following questions:
\textit{
    (1) Can we effectively apply NLP toolkits, such as classifiers, machine translation systems, and syntactic parsers, to visual representations of logographic writing systems?
    (2) Does this strategy allow for better transfer from pre-trained models and lead to better performance?
}
Additionally, as shown in \cref{fig:old_chinese}, many logographic languages have multiple partially processed representations, including artifact photographs, hand-copied lineart, Unicode, Latin transliteration, and normalization---we also aim to empirically investigate the extent to which various representations at each stage, including textual and visual modalities, facilitate effective fine-tuning of downstream NLP systems.

We have curated \textbf{LogogramNLP}, a benchmark consisting of four representative ancient logographic writing systems (Linear A, Egyptian hieroglyphic, Cuneiform, and Bamboo Script), along with annotations for fine-tuning and evaluating downstream NLP systems on three tasks, including three attribute classification tasks, machine translation, and dependency parsing. 

We conduct experiments on these languages and tasks with a suite of popular textual and visual encoding strategies.
Surprisingly, visual representations perform better than conventional text representations for some tasks (including machine translation), likely due to visual encoding allowing for better transfer from cross-lingual pre-training.
These results highlight the potential of visual representation processing, a novel approach to ancient language processing, which can be directly applied to a larger portion of existing data.

\begin{table}[!ht]
    \centering
    \resizebox{.45\textwidth}{!}{
        \begin{tabular}{@{}lrrrrr@{}}
            \toprule
            status         & LNA                                       & AKK                                       & EGY                                       & ZHO                                       & GRC                                      \\ \midrule
            deciphered     & \colorbox{red!70}{\makebox[0.8cm]{None}}  & \colorbox{blue!25}{\makebox[0.8cm]{Most}} & \colorbox{blue!25}{\makebox[0.8cm]{Most}} & \colorbox{blue!25}{\makebox[0.8cm]{Most}} & \colorbox{blue!40}{\makebox[0.8cm]{All}} \\
            differentiated & \colorbox{blue!25}{\makebox[0.8cm]{Most}} & \colorbox{blue!25}{\makebox[0.8cm]{Most}} & \colorbox{blue!25}{\makebox[0.8cm]{Most}} & \colorbox{blue!25}{\makebox[0.8cm]{Most}} & \colorbox{blue!40}{\makebox[0.8cm]{All}} \\
            encoded        & \colorbox{blue!25}{\makebox[0.8cm]{Most}} & \colorbox{blue!25}{\makebox[0.8cm]{Most}} & \colorbox{red!30}{\makebox[0.8cm]{Some}}  & \colorbox{red!30}{\makebox[0.8cm]{Some}}  & \colorbox{blue!40}{\makebox[0.8cm]{All}} \\
            Latinized      & \colorbox{blue!40}{\makebox[0.8cm]{All}}  & \colorbox{blue!40}{\makebox[0.8cm]{All}}  & \colorbox{blue!40}{\makebox[0.8cm]{All}}  & \colorbox{red!70}{\makebox[0.8cm]{None}}  & \colorbox{blue!40}{\makebox[0.8cm]{All}} \\
            \bottomrule
        \end{tabular}
    }
    \caption{Summary of the status of the ancient logographic languages presented in our paper. The status is measured from the perspective of paleography. We put Ancient Greek (GRC), a well-known ancient non-logographic language, here for comparison.}  \label{tb:status}
\end{table}

\begin{figure*}[t]
    \centering
    \includegraphics[width=\textwidth]{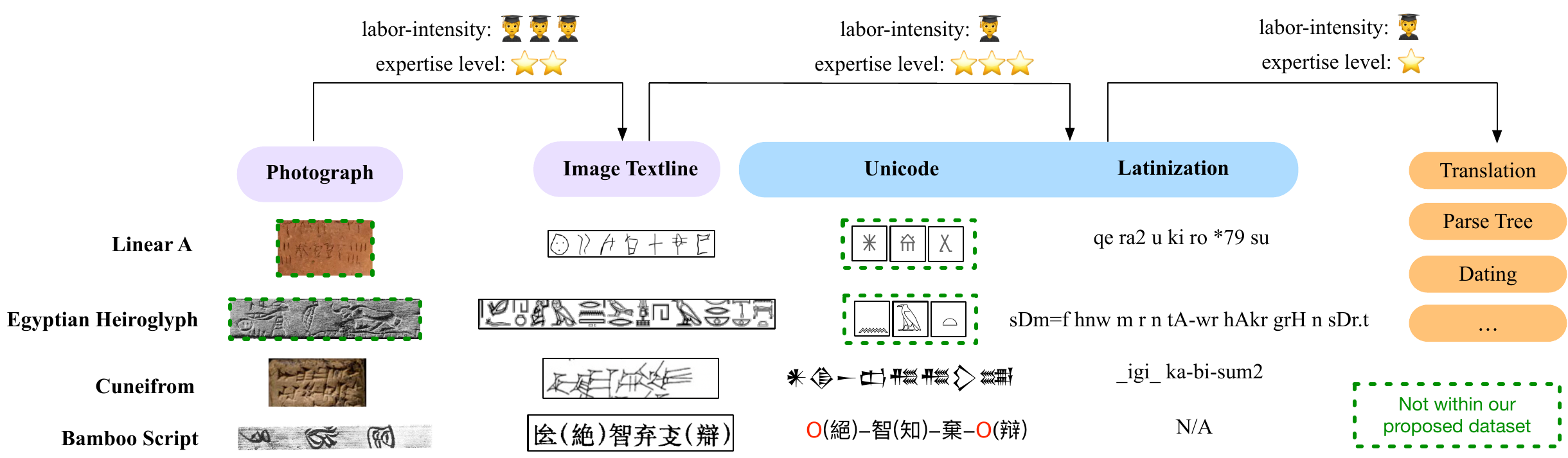}
    \caption{
        Example of four logographic languages with different representation formats.
        The arrow shows the typical processing flow of ancient languages by humanists.
        The workload and expertise required to transcribe the text from images is even greater than that of downstream tasks such as machine translation. The red circle \color{red}{O} \color{black} (in Bamboo Script) indicates the character is not digitized as Unicode yet.
        Green dashed boxes note that Unicode exists for Egyptian hieroglyphics and Linear A, but the alignment to documents is unavailable; the same goes for Egyptian and Linear A photographs.
    }
    \label{fig:overview}
\end{figure*}

\section{Dataset: Languages, Tasks and Challenges}
\label{sec:data-and-challenge}

Our benchmark consists of four representative ancient languages---Linear A, Egyptian hieroglyphic, Cuneiform, and Bamboo script (\cref{sec:logographic-languages}).\footnote{Bamboo scripts usually combine Seal scripts and Clerical scripts.}
Each language is associated with a unique writing system and unique challenges.
We refer the readers to \cref{sec:datacollection} for data collection and cleaning details.
Our benchmark covers three tasks: machine translation, dependency parsing, and attribute classification (\cref{sec:tasks}).

\subsection{Logographic Languages}
\label{sec:logographic-languages}

A major characteristic of logographic languages is that the size of symbol inventories is significantly larger than that in alphabetic languages such as Ancient Greek (24 letters) or Modern English (26 letters).
A summary of different representations of the languages of our interest is shown in \cref{fig:overview}, and \cref{tb:status} summarizes the current status of each language.

\paragraph{Linear A.}
Linear A is an undeciphered language used by the Minoan at Crete and is believed to be not related to ancient Greek.
Scholars have differentiated the glyphs and carefully hand-copied them into linearts.
We collected a dataset of 772 tablets (i.e., manually drawing) from SigLA.\footnote{\url{https://sigla.phis.me/browse.html}}
Each tablet also has a separable glyph with annotated Unicode.

\paragraph{Akkadian (Cuneiform).}
CuneiML \citep{chen2023cuneiml} is a dataset that contains 36k entries of cuneiform tablets.
Each tablet consists of Unicode Cuneiform, lineart, and transliteration.
We also use the Akkadian Universal Dependencies (UD) dataset \citep{luukko-etal-2020-akkadian}, which contains 1,845 sentences with dependency annotations.
Since the UD annotation of Akkadian only keeps the normalization form of the language, we obtain the Unicode by running a dynamic programming-based matching algorithm.

\paragraph{Ancient Egyptian (Hieroglyph).}
We segmented the St Andrews Corpus \citep{nederhof2015ocr}\footnote{\url{https://mjn.host.cs.st-andrews.ac.uk/egyptian/texts/corpus/pdf/}} using a rule-based segmenter, and obtained 891 examples of parallel data.
Additionally, we collected data from the Thot Sign List (TSL; English translation)\footnote{\url{https://thotsignlist.org/}} and BBAW (German translation)\footnote{\url{https://aaew.bbaw.de/tla/servlet/TlaLogin}} for 2,337 and 100,736 samples of parallel data, respectively. However, the transliteration standards differ among these three sources of data, and BBAW does not include hieroglyph image features. Therefore, we only used TSL's data.

\paragraph{Old Chinese (Bamboo script).}
We collected 13,770 pieces of bamboo slips from Kaom,\footnote{\url{http://www.kaom.net}} which come with the photograph of each line of the text.
The Baoshan collection covers three genres: Wenshu (Document), Zhanbu (Divine), and Book.
The Guodian collection contains parallel data translated into modern Chinese.
The vocabulary size is 1,303.
Notably, about $40\%$ of the characters do not have a Unicode codepoint and are, therefore, represented as place-holder triangles or circles.
This dataset does not come with human-labeled latinization due to the lack of transliteration standards.

\subsubsection{Visual Representations}
\label{sec:visual-feature-linearization}
Since ancient scripts did not consistently adhere to a left-to-right writing order, breaking down multi-line documents into images of single-line text is nontrivial.
These historical data, therefore, need additional processing to be machine-readable.
\cref{fig:glyph} shows examples of different processing strategies.
We summarize the approaches we used in building the dataset as follows:

\begin{enumerate}[leftmargin=*,topsep=1pt,noitemsep]
    \item \textbf{Raw image (no processing)}: the raw images are already manually labeled and cut into text lines of images, and no extra processing is required.
    \item \textbf{Montage}: we generate a row of thumbnails of each glyph using the montage tool in ImageMagick.\footnote{\url{https://imagemagick.org}}
          This strategy is used for Linear A, as the original texts are written on a stone tablet, and scholars have not determined the reading ordering of this unknown script.
    \item \textbf{Digital rendering}: we digitally render the text using computer fonts when the language is already encoded in Unicode.
          Given that most ancient logographic scripts are still undergoing the digitization process, this option is currently unavailable except for Cuneiform.
\end{enumerate}

\subsubsection{Textual Representations}
The processing of textual features for ancient logographic scripts also requires special attention.
Unlike modern languages, ancient logographic writing systems can have multiple latinization standards or lack universally agreed-upon transcription standards.
For example, the cuneiform parsing data is not in standard transliteration (ATF)\footnote{ATF is a format used to represent cuneiform text.
    More details can be found at \url{http://oracc.ub.uni-muenchen.de/doc/help/}
} form, but rather, in the UD normalized form.
This mismatch introduces extra difficulty to downstream tasks, especially in low-resource settings.

A similar issue also exists for Old Chinese: most ancient characters do not even exist in the current Unicode alphabet.
While we may find some modern Chinese characters that look similar to the ancient glyphs, they are usually not identical, and such a representation loses information from the original text.

\begin{figure*}[t]
    \centering
    \includegraphics[width=\linewidth]{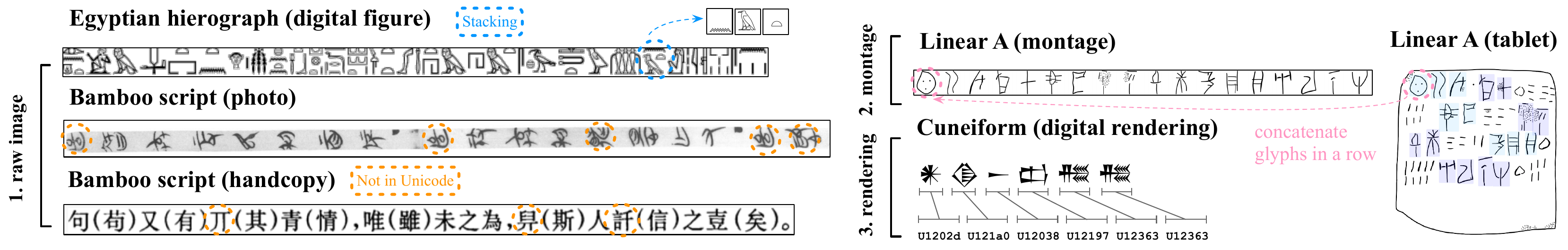}
    \caption{Image features of four ancient writing systems. (1) Egyptian hieroglyphs and Bamboo scripts are already manually segmented into images of lines. In the handcopy version of the Bamboo script, the word within parentheses indicates the corresponding modern Chinese glyph. Although both the Egyptian and Bamboo script images appear to be in a digital font, they are only accessible as images without underlying codepoint mappings to Unicode. (2) Linear A tablets are believed to be written in  horizontal lines running from left to right \cite{salgarella2020aegean}; therefore, we use the montage concatenation of each glyph as the representation. (3) We digitally render Cuneiform Unicode using computer font as the visual representation.  
    }\label{fig:glyph}
\end{figure*}

For Egyptian hieroglyphs, most characters are encoded in Unicode, but there is no standard encoding for ``stacking'' multiple glyphs vertically (\cref{fig:glyph}).
Therefore, we do not include the Unicode text for our ancient Egyptian data as they are not available.

\subsection{Tasks}
\label{sec:tasks}
Our benchmark covers three tasks (\cref{tb:tasksum}): translation, dependency parsing, and attribute classification.
The model performance on these tasks reflects various aspects of ancient language understanding.
To better understand the information loss when using a pipeline approach, we also report performance using this method: predicting the transliteration first and using the noisy predicted transliteration for downstream tasks.

\paragraph{Machine translation.}
The task is to translate the ancient languages, represented by either text or images, into modern languages, such as English.
In all of our experiments, we translate ancient languages into English.

\paragraph{Dependency parsing.}
Given a sentence in the ancient language, the task is to predict the dependency parse tree \citep{tesniere1959elements} of the sentence.
In the dependency parse tree, the parent of each word is its grammatical head.

\paragraph{Attribute classification.} The task is to predict an attribute of the given artifact, for example, provenience (found place), time period, or genre.

\section{Methods}

In this section, we will describe feature encoding methods (\cref{sec:feat_enc}) for both visual and textual inputs, as well as task-specific layers (\cref{sec:task_spec}) for each task we consider.

\subsection{Feature Encoding}
\label{sec:feat_enc}

NLP for Low-resource languages has benefitted a lot from pre-trained models.
However, modern pre-trained models do not cover the character inventories of the considered ancient logographic languages.
To overcome this shortage, we summarize solutions to the problem into four categories and describe them as follows.

\paragraph{Extending vocabulary.}
In this line of approach \citep{wang2020extending,imamura2022extending}, the vocabulary is extended by adding the unseen tokens.
The embeddings of new tokens can be either initialized randomly or calculated by a function.
In the fine-tuning stage, the embeddings of new tokens are updated together with the rest of the model.

\paragraph{Latin transliteration as a proxy.}
The majority of past work on cross-lingual transfer has focused on using Latin transliteration as the proxy to transfer knowledge from high-resource to low-resource languages \cite{pires2019multilingual,amrhein-sennrich-2020-romanization}.
Following this line of work, we input latinization representations to mBERT \citep{devlin2018bert} to obtain the embeddings of the ancient languages.

\paragraph{Tokenization-free.}
The idea of the tokenization-free approach is to view tokens as a sequence of bytes and directly operate on UTF-8 codepoints without an extra mapping step.
As representative models, ByT5 \citep{xue2022byt5} and CANINE \citep{10.1162/tacl_a_00448} use Unicode encoding of a string to resolve the cross-lingual out-of-vocabulary issues.
This work uses ByT5 for machine translation and CANINE for classification.

\paragraph{Pixel Encoder for Text.}
Recently, there has been a novel approach \citep{rust-etal-2023-pixel} that aims to resolve the disjoint-character-set problem by rendering text into images and then applying a standard image encoder, such as the Vision Transformer with Masked Autoencoder (ViT-MAE) \citep{he2022masked}, to encode the features.
In this work, we use PIXEL \citep{rust-etal-2023-pixel}, a pixel-based language model pre-trained on the Common Crawl dataset with a masked image modeling objective, to encode the visual text lines for ancient languages.
Additionally, we use PIXEL-MT \citep{salesky-etal-2023-multilingual}, a pixel-based machine translation model pre-trained on 59 languages, for the machine translation task.

\paragraph{Full Document Image Encoding.}
When the images of ancient artifacts are available (e.g., for Linear A and Cuneiform), we can encode the full-document images directly.
We use ResNet-50 \citep{he2016deep} as the backbone model for full-document image inputs.

\subsection{Task-Specific Layers}
\label{sec:task_spec}

\paragraph{Machine translation.}
After encoding the input to vectors, machine translation requires a decoder to generate sequential outputs.
Encoder-decoder models, such as T5 \citep{raffel2020exploring}, ByT5, PIXEL-MT, and BPE-MT \cite{salesky-etal-2023-multilingual}, use 3/6/12 layers of Transformer blocks as the decoders.
For Encoder-only models, such as (m)BERT or PIXEL, we attach a GPT2 model \citep{radford2019language} as the decoder to produce sequential output.
Among the aforementioned models, T5, ByT5, and PIXEL are pre-trained on large-scale text corpora such as the Common Crawl; PIXEL-MT and BPE-MT are pre-trained on 1.5M pairs of sentences of 59 modern languages; PIXEL-MT is an encoder-decoder model with a 6-layer Transformer encoder and a 4-layer Transformer decoder.

\paragraph{Classification.}
We attach a two-layer ReLU-activated perceptron (MLP) with a hidden size of 512 to the encoder for all classification tasks.
The MLP outputs the predicted distribution over the candidate classes.

\paragraph{Dependency Parsing.}
After encoding, we use the deep bi-affine parser \citep{dozat2017deep} for dependency parsing, which assigns a score to each possible dependency arc between two words.
We use the minimum spanning tree (MST) algorithm during inference to find the best dependency tree for each sentence. 

\section{Experiments and Analysis}

We describe our general model fine-tuning approach in \cref{sec:general_setup} and analyze model performance on the aforementioned tasks in the succeeding subsections.

\begin{table}[t]
  \centering
  \resizebox{\columnwidth}{!}{%
    \begin{tabular}{@{}ccccc@{}}
      \toprule
      \textbf{Task}  & \textbf{Model} & \textbf{BSZ} & \textbf{Steps} & \textbf{LR} \\ \midrule
      translation    & visual         & 64           & 30,000         & 5e-4        \\
      translation    & textual        & 56           & 30,000         & 5e-4        \\
      translation    & byT5           & 64           & 100,000        & 1e-3        \\
      classification & visual/textual & 256          & 30,000         & 5e-4        \\
      parsing        & visual/textual & 256          & 1,000          & 8e-5        \\
      \bottomrule
    \end{tabular}
  }
  \caption{Hyperparameter configuration. Note that, byT5 is particularly hard to converge compared to other transformer-based models. For the parsing task, due to the low-resource nature of the parsing data, 1,000 steps are sufficient to achieve model convergence.
  }\label{tb:config}
\end{table}

\begin{table*}[!ht]
  \centering
\begin{subtable}{\textwidth}
\centering
  \caption{Machine translation (BLEU score)}
\resizebox{0.8\linewidth}{!}{%
\begin{tabular}{@{}rlllcccrccc@{}}
    \toprule
    \multirow{2}{*}{\textbf{Modality}} & \multirow{2}{*}{\textbf{Tokenization}} & \multirow{2}{*}{\textbf{Input}}  & \multirow{2}{*}{\textbf{Model}}         & \multicolumn{2}{c}{\textbf{Pre-trained?}} &\multicolumn{3}{c}{\textbf{Source Language}}                                              \\
    \cmidrule(lr){5-6} \cmidrule(lr){7-9}
          &    &      &       & MLM                                      & MT                                                                           & \textbf{EGY}                              & \textbf{AKK}      & \textbf{ZHO}     \\
    \midrule
    &&  \multicolumn{4}{r}{\texttt{Dataset size (\# lines)}} & \texttt{2,337} & \texttt{8,056} & \texttt{500} \\\midrule
    \textbf{Visual} & token-free & textline & PIXEL + GPT2$^1$             & \Checkmark                                & \xmark                                                                  & 2.83                                      & 7.51              & 1.14                              \\
    \textbf{Visual} & token-free  & textline & PIXEL-MT                  & \xmark                                    & \Checkmark                                                           & \textbf{29.16}                            & \textbf{44.15}    & \textbf{5.45}                     \\\midrule

    \textbf{Textual}  & BPE \textit{w/ ext vocab}$^2$ & Unicode & T5  & \Checkmark                                & \xmark                                                              & n/a                                       & 12.42             & 0.28                              \\
     \textbf{Textual} & byte-level  & Unicode  & ByT5                      & \Checkmark                                & \xmark                                                           & n/a                                       & 4.51              & 0.53                              \\
         \textbf{Textual} & char-level & Unicode   & Conv-s2s           & \xmark                                    & \xmark                                                              & -                                         & 36.52$^{*}$              & -                                 \\

         \textbf{Textual} & BPE & Unicode & BPE-MT                    & \xmark                                    & \Checkmark                                              & \underline{23.26}                         & 36.18             & \underline{1.32}                  \\
    \textbf{Textual} & BPE  & Latin  & T5                 & \Checkmark                                & \xmark                                                              & 21.18                                     & 10.67             & n/a                               \\
     \textbf{Textual} & char-level & Latin  & Conv-s2s                & \xmark                                    & \xmark                                                               & -                                         & \underline{37.47}$^{*}$ & -                                 \\
    \bottomrule
  \end{tabular}\label{tb:MT}
  } 
\end{subtable}
\begin{subtable}{\linewidth}
    \centering
  \caption{Attribute prediction (F$_1$ accuracy)}
   \resizebox{0.8\linewidth}{!}{%
\begin{tabular}{@{}rlllcccccc@{}}
    \toprule
    \multicolumn{1}{c}{\textbf{Modality}}             &
    \multicolumn{1}{c}{\textbf{Tokenization}} &
    \multicolumn{1}{l}{\textbf{Input}}              &
    \multicolumn{1}{l}{\textbf{Model}}              &
    \multicolumn{1}{c}{\textbf{LNA}}                &
    \multicolumn{3}{c}{\textbf{AKK}}                &
    \multicolumn{1}{c}{\textbf{EGY}}                &
    \multicolumn{1}{c}{\textbf{ZHO}}                                                                                                                                                                                                                          \\\cmidrule(lr){5-5} \cmidrule(lr){6-8} \cmidrule(lr){9-9} \cmidrule(lr){10-10}
                                                    &       &   &                            & \multicolumn{1}{c}{geo} & \multicolumn{1}{c}{time} & \multicolumn{1}{c}{genre} & \multicolumn{1}{c}{geo} & \multicolumn{1}{c}{time} & \multicolumn{1}{c}{genre} \\\midrule
  &  \multicolumn{3}{r}{ \texttt{Number of classes}} & 7        & 16                         & 12                      & 24                       & 14                        & 3
    \\
  &  \multicolumn{3}{r}{\texttt{Dataset size (\# examples)}}       & 772      & 36,454                     & 36,454                  & 36,454                   & 1,320                     & 302                                                                            \\
    \midrule
                          &                        &          & Majority                   & 14.28                   & 6.25                     & 8.33                      & 4.17                    & 7.14                     & 33.33                     \\ \midrule
    \textbf{Visual}                &  token-free                 & photo    & ResNet                     & 8.24                    & 75.02                    & 45.45                     & 62.99                   & \multicolumn{1}{c}{n/a}  & \multicolumn{1}{c}{n/a}   \\
    \textbf{Visual}             &  token-free          & textline & PIXEL                      & 16.56                   & 72.91                    & 50.84                     & 61.44                   & 16.24                    & 52.17                     \\\midrule
    \textbf{Textual}        &   BPE \textit{w/ ext vocab}$^{2}$    & Unicode  & BERT                       & \multicolumn{1}{c}{n/a} & $0^{**}$                  & $0^{**}$                   & $0^{**}$                 & \multicolumn{1}{c}{n/a}  & 74.85                     \\
    \textbf{Textual}            &   BPE       & Unicode  & BERT  & \multicolumn{1}{c}{n/a} & 72.40                    & 50.85                     & 63.70                   & \multicolumn{1}{c}{n/a}  & \underline{90.30 }        \\
    \textbf{Textual}       & byte-level            & Unicode  & CANINE                     & \multicolumn{1}{c}{n/a} & \underline{82.83}        & 47.88                     & 56.42                   & \multicolumn{1}{c}{n/a}  & \textbf{96.43}            \\
    \textbf{Textual}       &  BPE                      & Latin    & BERT                       & \underline{32.92}       & 80.91                    & \underline{53.45}         & \underline{65.10}       & \underline{34.71}        & \multicolumn{1}{c}{n/a}   \\
    \textbf{Textual}           &  BPE             & Latin    & mBERT                      & \textbf{50.52}          & \textbf{83.08}           & \textbf{56.71}            & \textbf{66.33}          & \textbf{36.25}           & \multicolumn{1}{c}{n/a}   \\
    \bottomrule
  \end{tabular}\label{tb:attr_pred}
  } %
  \end{subtable}

  \caption{
    (a) Results on machine translation (from each of the source languages to English), in terms of BLEU scores.
    MLM denotes models pretrained on unsupervised data with the masked language model (MLM) loss, while MT denotes models pretrained with supervised parallel data (TED59). (b) Macro F$_1$ scores for attribute prediction. \\
    ${^*}$: numbers taken from \citet{gutherz2023translating}, where their models are trained from scratch, i.e., without pretraining.
    ${^{**}}$: The character set is 100\% disjointed without extending the vocabulary of the model, resulting in zero F$_1$ scores.
    ${^1}$: This model is trained using PIXEL as the encoder and GPT2 as the decoder, with linear projection layers to convert the final layer of PIXEL into a prefix input for GPT2.
    ${^2}$: This model is the only one experiencing out-of-vocabulary (OOV) issues with Unicode input. To address this, we extended the vocabulary with random initialization.
    \textbf{n/a}: indicates the representation of a specific language does not exist in our benchmark.
  }
\end{table*}

 

\subsection{General Experimental Setup}\label{sec:general_setup}

We use the Huggingface Transformers library \citep{wolf-etal-2020-transformers} in all experiments, except for machine translation, where we use the PIXEL-MT and BPE-MT models.\footnote{The prefix PIXEL or BPE also indicates the type of input representation the model uses.}
We modified code and model checkpoints provided by \citet{salesky-etal-2023-multilingual} based on fairseq \citep{ott2019fairseq} for the two exceptions.

We use Adam \citep{kingma-ba-2015-adam} as the optimizer for all models, with an initial learning rate specified in \cref{tb:config}.
We use early stopping when the validation loss fails to improve for ten evaluation intervals (1000 iteration per interval).
For data without a standard test set, we run a fixed number of training iterations and report the performance on the validation set after the last iteration.
All experiments are conducted on an NVIDIA-RTX A6000 GPU, and the training time ranges from 2 minutes to 50 hours, depending on the nature of the task and the size of the datasets.
Unless otherwise specified, all parameters, including those in pre-trained models, are trainable without freezing.
We summarize other configurations in \cref{tb:config}.

\subsection{Machine Translation}
We compare the performance of the models on machine translation, where we translate ancient Egyptian (EGY), Akkadian (AKK), and Old Chinese (ZHO) into English (\cref{tb:MT}).
We find that the PIXEL-MT model consistently achieves the best BLEU score across the three languages, outperforming the second-best method by a large margin.

Models with pre-training do not always outperform those trained from scratch \citep{gutherz2023translating}.
We find that all models that take textual (Unicode or latinized) input achieve worse performance than models trained from scratch with the same type of textual input, suggesting that the lack of overlap in symbol inventories poses a serious problem for cross-lingual transfer learning. 
Our results indicate that choosing the correct input format is crucial to achieving the full advantage of pre-training.

In addition, the PIXEL-MT model, pre-trained on paired data in modern languages (TED59), significantly outperforms PIXEL + GPT2 (pre-trained with masked language modeling) across the board.
Another model, BERT-MT, which is further pre-trained on the same parallel text (TED59) with BERT initialization, also achieves comparable performance.
These results emphasize the importance of pre-training on modern paired data, empirically suggesting that the PIXEL encoder with parallel text pretraining is an effective combination for ancient logographic language translation.

\paragraph{Qualitative analysis.}
\begin{figure}[h]
  \centering
  \includegraphics[width=0.5\textwidth]{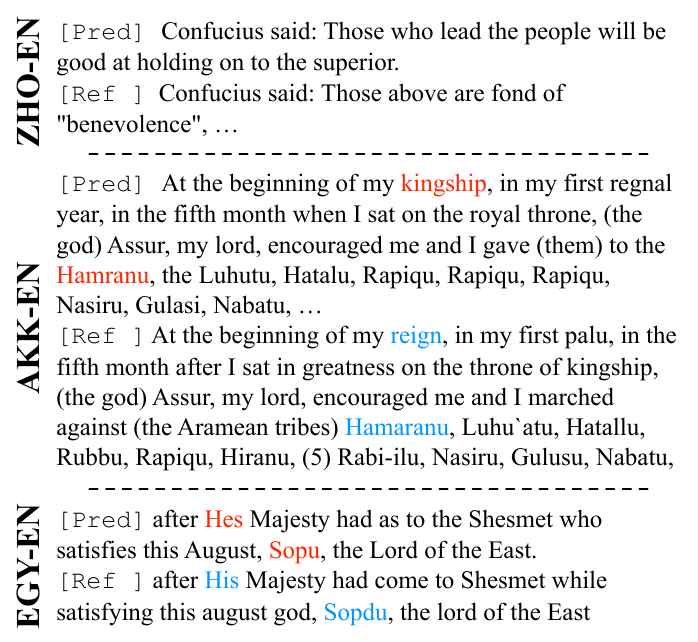}
  \caption{
    Case study for machine translation using the PIXEL-MT model.
    Notably, there are many spelling errors in the predictions, particularly with uncommon named entities.
  }\label{fig:translation_case}
\end{figure}

As shown in \cref{fig:translation_case}, the low BLEU scores for ZHO-EN translation is a result of the translation model failing to capture the meaning of the input, instead focusing on repeated formatting queues: e.g., ``\texttt{Confucius said: Those who...}''
Indeed, given that the topical domain of the ZHO-EN translation data is philosophical writing, achieving an accurate translation would be challenging even with a much larger set of parallel translations.
For AKK-EN, we found that the overall quality to be quite good, despite the fact that errors in translating named entities appear more often than in standard MT tasks.
This case study suggests that translation performance could improve further if we training using a custom target language (English) vocabulary.
We also show more generated examples from the \texttt{PIXEL-MT} model in \cref{sec:more_trans_examples}.

\subsection{Attribute Classification}

\cref{tb:attr_pred} summarizes the performance of attribute classification with different features and models.
As expected, the image features can work fairly well for some of these attribute classification tasks as many of the relevant features are visual (e.g., for time and location); but, are not generally as effective as textual input representations.
By comparing BERT with latinized input and CANINE on Unicode, we find that when both accurate latinization and Unicode representations are available, latinization is the most informative feature---with the exception of time period classification for Akkadian.
This exception is aligned with our understanding of Akkadian, as different Cuneiform characters are used across different time periods. Thus, in this case, Unicode can provide more clues for determining the time period of a sample. Note that the label distribution is not balanced for most ancient language attribution tasks. For more details, refer to \citet{cuneidate}.

\subsection{Dependency Parsing}
\begin{table}[t]
\centering
\resizebox{0.95\columnwidth}{!}{%
\begin{tabular}{@{}lllrr@{}}
\toprule
 \textbf{Modality} & \textbf{Model}  & \textbf{Input} & \textbf{RIAO} & \textbf{MCONG} \\ \midrule
 
 \multicolumn{3}{r}{\texttt{Dataset Size (\# tokens)}} & \texttt{5k} & \texttt{130k} \\\midrule
  Visual & PIXEL & Image    & \textbf{92.74}                  &       \textbf{85.22}            \\
   Textual & BERT & Latin   &           92.13               &         83.88           \\ 
     \bottomrule
\end{tabular}
}
\caption{Dependency parsing result on Akkadian (evaluated on the UD corpora RIAO and MCONG), in terms of labeled attachment scores (LAS).  Note that the number of tokens are reported.
}\label{tb:depar}
\end{table}
We compare the dependency parsing performance of models with visual and textual encoders (\cref{tb:depar}).\footnote{We only conduct experiments on Akkadian since it is the only language with off-the-shelf dependency annotations.}
While all models achieve quite high parsing accuracy, we find that models with visual encoders perform the best on both investigated corpora (RIAO and MCONG).
During training, models taking visual input generally converge faster than their textual counterparts, which is in line with prior work \citep{salesky-etal-2023-multilingual} that uses visual features for machine translation.

\section{Ablation Study on OCR and Image Quality}
As mentioned earlier, the majority of data from ancient times remain in the form of photographs.
We first closely examine two different visual input representations for the ZHO-EN translation task, \textit{handcopied figure} and \textit{photograph} (\cref{sec:handcopy}).
Next, we examine OCR performance on ancient logographic languages to gain better understanding of this bottleneck for current NLP pipelines (\cref{sec:ocr}).

\subsection{Handcopy v.s. Raw Image}
\label{sec:handcopy}
\begin{table}[h]
  \centering
  \resizebox{0.6\columnwidth}{!}{
  \begin{tabular}{@{}lr@{}}
    \toprule
    \textbf{Input representation} & \textbf{BLEU} \\ \midrule
    photograph                    & 2.09          \\
    handcopied figure             & 5.45          \\
    \bottomrule
  \end{tabular}
  }
  \caption{
    Performance on ZHO-EN translation using the PIXEL-MT model with different visual input features.
  }
  \label{tb:vis_translation}
\end{table}

For the ZHO-EN translation data, we have access to both photographs of the bamboo slips and handcopied textline figures (see the Bamboo script example in \cref{fig:glyph} for reference).
As shown in \cref{tb:vis_translation}, the quality of the visual features significantly influences the translation accuracy---translations derived from photographs yield a low BLEU score of 2.09, whereas handcopied figures, which typically provide clearer and more consistent visual data, result in a higher BLEU score of 5.45.
This result suggests that for models that perform implicit OCR as part of the translation process, the clarity of the source material is paramount.

\subsection{Text Recognition Study}
\label{sec:ocr}

We simplify the task of transcribing ancient texts by starting with lines of text that have been accurately segmented. For datasets that include glyph-level annotations, we employ glyph classification to recognize the text. Details on models and configuration of \textit{line-level OCR} and \textit{glyph classification} can be found in Appendix \ref{ch:detail_OCR}.

\begin{table}[h]
  \centering
  \resizebox{0.9\columnwidth}{!}{%
    \begin{tabular}{@{}rrrrrrr@{}}
      \toprule
      \textbf{Method} & \textbf{Output} & \textbf{LNA} & \textbf{EGY} & \textbf{AKK} & \textbf{ZHO} \\ \midrule
      OCR             & Unicode         & 57.17        & N/A          & 5.72         & 71.85        \\
      OCR             & Latin           & 63.44        & 65.88        & 21.98        & N/A          \\
      \bottomrule
    \end{tabular}
  }
  \caption{
    Line-level OCR results with the best validation character error rate (CER) reported.
    The study includes various writing systems using Kraken trained from scratch on segmented text lines.
    N/A: either the Unicode or Latin version of the text is not available. 
  }
  \label{tb:OCR_main}
\end{table}

\begin{figure}[h]
  \centering
  \includegraphics[width=0.95\columnwidth]{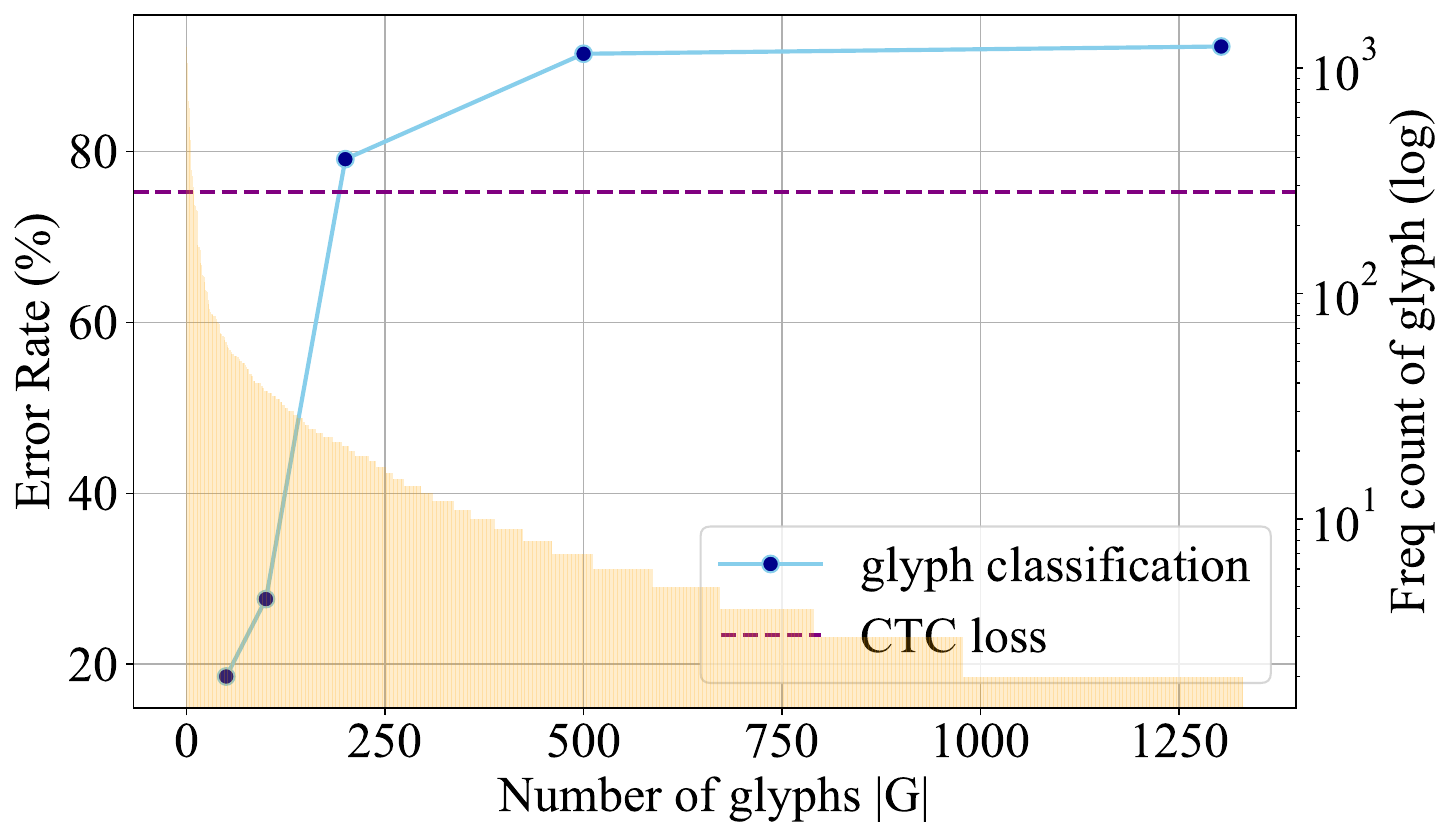}
  \caption{
    Glyph classification on Old Chinese (ZHO). \textbf{Left axis}: we plot the error rate of glyph classification. The data point at |G| = 50 shows the classification error calculated using the top 50 most frequent glyphs in the dataset. The purple horizontal line (71.85\%) represents the line-level text recognition CER for ZHO, provided for reference. \textbf{Right axis}: The frequency count (in orange bars) of each glyph in the dataset. Note that the counts are in logarithmic scale, illustrating the long tail distribution of glyph counts.
  }\label{fig:OCR_glyph}
\end{figure}

\paragraph{Results.}
The line-level OCR performance for the four languages is presented in Table \ref{tb:OCR_main}.
When comparing digital renderings of text to handwritten samples, it is evident that Old Chinese (ZHO) achieves a CER of 71.85, while Linear A has a CER of 57.17.
As shown in Figure \ref{fig:OCR_glyph}, glyph classification for ZHO is approximately 20\% less accurate than line-level OCR, indicating that contextual features significantly aid in recognizing glyphs.
Furthermore, there is a rapid increase in error rate as the number of glyphs increases, highlighting the intrinsic challenge of processing logographic languages, which typically have a large symbol inventories, and their frequency distribution often follows a long-tail pattern (see the orange bars in Figure  \ref{fig:OCR_glyph}). Therefore, developing robust visual models that can effectively leverage visual features is crucial for improving NLP on ancient logographic languages.

\section{Related work}
Because ancient languages are often low-resource, they present challenges that are closely related to other domains of NLP, such as low-resource machine learning and multi-lingual transfer learning.
Recent work has explored the application of NLP techniques to ancient languages from the following perspectives:

\paragraph{Multilingual transfer learning and disjoint character sets.} \citet{muller2020unseen} studied hard-to-process living languages such as Uyghur, and reported that a non-contextual baseline outperforms all pre-trained LM-based methods. Ancient languages also face the same problem, with even less data available.
A major challenge that is mostly specific to ancient logographic languages, however, is the almost non-existent overlap of their symbol inventories with those of high-resource languages.

\paragraph{Visual representation of languages.} Recently, several works have studied language processing based on images of text. \citet{rust-etal-2023-pixel} pre-trained a masked language model on digitally rendered text and achieved comparable performance with text-based pre-training strategies on downstream tasks. \citet{salesky-etal-2023-multilingual} found that a multi-lingual translation system with pixel inputs was able to outperform its textual counterpart.

\paragraph{Machine learning for ancient languages.}
\citet{sommerschield2023machine} surveyed the status of pipelines for ancient language processing.
Notably, the study concludes that applying machine learning methods to ancient languages is bottlenecked by the cost of digitization and transcription.
According to the Missing Scripts Project,\footnote{\url{https://worldswritingsystems.org/}} only 73 of 136 dead writing systems are encoded in Unicode.
Ancient languages, such as Ancient Greek or Latin \citep{bamman2020latin}, benefit greatly from multilingual pre-training techniques, such as mBERT, XLM-R \citep{conneau2020unsupervised}, and BLOOM \citep{scao2022bloom}.
The applicability of these techniques is limited when it comes to languages that were historically written in obsolete or extinct writing systems---for instance, languages like Sumerian and Elamite were recorded in Cuneiform script and ancient Chinese was inscribed on oracle bone or bamboo.
However, observations by existing work support the potential utility of visual processing pipelines for ancient languages.

\paragraph{Logographic writing systems.} Logography typically denotes a writing system in which each glyph represents a semantic value rather than a phonetic one, however, all the languages studied in our paper have at least some phonetic component based on the rebus principle. 
This paper emphasizes \textit{ancient} logographies that (i) possess extensive glyph inventories; (ii) have feature glyphs with ambiguous transliterations or functional uses; and (iii) are low-resource with much of data remaining in photo formats \citep{caplice1991introduction,allen2000middle,woodard2004cambridge}.
Existing research on logographic languages has predominantly focused on those well-resourced and still in use, such as Modern Chinese \citep{zhang-komachi-2018-neural,si-etal-2023-sub}, or used data that has already been carefully transcribed into Latin or annotated with extra semantic information \citep{wiesenbach2019multi,gutherz2023translating,jiang2024finding}. 
Our paper aims to address the gap in resources (by proposing new data) and methodologies (by adapting visual-only approaches) for encoding and analyzing ancient logographic languages, leading to more comprehensive understanding of historical linguistic landscapes.

\section{Conclusion}

By comparing the results on four representative languages on three downstream tasks, we demonstrated the challenges faced in applying natural language processing techniques to ancient logographic writing systems. Our experiments demonstrate, however, that encoding more readily available visual representations of language artifacts can be a successful strategy for downstream processing tasks.

\section{Limitations}

\paragraph{More discussion on ancient logographic languages.} Due to page limits, we do not discuss ancient logographic languages in a critical way. Technically, there are no logographic languages, only languages written in logographic writing systems (aka \textit{logography}) \cite{gorman-sproat-2023-myths}. In this paper, we use the term ``logographic languages'' to denote languages that are quite different from those with alphabetic writing systems especially when we tried to apply NLP toolkits for computational paleography. As mentioned in the related work section, these languages feature glyphs that have multiple transliterations or functional uses. In other words, these languages are homophonous or a glyph can be used as a phonetic value or semantic value. Therefore, the boundaries between logographic and phonographic is not sharply separated.

\paragraph{Including more logographic writing systems.} We selected the four languages because we would like to include at least one language from early civilization in Ancient China, Ancient Egypt, Indus Valley Civilization, Mesoamerica, Mesopotamia and Minoan Civilization \cite{woodard2004cambridge}. However, we fail to include Mayan hieroglyphs (Mesoamerica) and Oracle Bone script. 
However, Mayan is excluded because the collection\footnote{The Maya Hieroglyphic Text and Image Archive: \url{https://digitale-sammlungen.ulb.uni-bonn.de/maya}} is still working in process. Oracle Bone script is primarily omitted due to copyright issues. 

\paragraph{Textline images.}
Most ancient languages remain as full-document images. 
In this paper, we use digitally rendered text as a surrogate visual feature for Akkadian.
In reality, much of Cuneiform data is still in hand copies or in photo format.
In the future, we look to conduct apples-to-apples comparisons for all languages once the line segmentation annotations become available.

\paragraph{Annotation quality and quantity.} The study of ancient languages is constantly evolving; humanities scholars have not agreed on explanations, transliterations, or even the distinctions between certain glyphs or periods. We try our best to carefully annotated the data without bias; however, future editions of the benchmark are needed as things change all the time. A collective platform to correct errors and make more data available should be considered for future development.

\paragraph{Label imbalance.} 
The classification task in our benchmark is label imbalanced. 
This is known to be a major issue for all machine learning tasks related to the ancient world \citep{sommerschield2023machine,cuneidate}.

\section*{Acknowledgements}

We thank Professor Wenbo Chen from the Department of Humanities at Central South University, China, for his advice on Old Chinese data collection and explanation.
We thank Professor Edward Kelting from the Department of Literature at UC San Diego for his advice on Ancient Egyptian data collection and explanation.
We thank Jerry You from \url{http://www.ccamc.org/} for his help on Unicode and data processing for ancient languages.

We thank Elizabeth Salesky for her guidance in setting up cross-lingual machine translation experiments for ancient languages using both PIXEL and BPE encoders.
We thank Chenghao Xiao for the help setting up the PIXEL + GPT 2 experiment.

We thank Kyle Gorman,  Alexander Gutkin and Richard Sproat for their inspiring work \cite{sproat2021taxonomy,gorman-sproat-2023-myths}, which has significantly contributed to our understanding of logographic writing systems from a computational perspective.

We thank Nikita Srivatsan, Nikolai Vogler, Ke Chen, Daniel Spokoyny, David Smith, and the anonymous ARR reviewers for their insightful feedback on the paper draft. This work was partially supported by the NSF under grants 2146151 and 2200333.

\bibliography{anthology,custom}
\bibliographystyle{acl_natbib}

\appendix

\section{Dataset collection}
\label{sec:datacollection}

\subsection{Linear A}

\subsubsection{Attribute classification}
Linear A is a logo-syllabic, undeciphered writing system that was used in Bronze Age Greece (ca. 1800-1450BCE). We crawled a dataset of 772 tablets from the publicly available  SigLA database \footnote{\url{https://sigla.phis.me/browse.html}}. The metadata includes find-place, tablet dimensions, total number of signs and possible transliteration. For the attribute classification task, we use find-place as the target class, with the following classes \texttt{['Arkhanes', 'Gournia', 'Haghia Triada', 'Haghios Stephanos', 'Kea', 'Khania', 'Knossos', 'Kythera', 'Malia', 'Mallia', 'Melos', 'Mycenae', 'Papoura', 'Phaistos', 'Psykhro', 'Pyrgos', 'Syme', 'Tylissos', 'Zakros']}. We only keep classes whose tablets are no less than 10, which results in only keeping 7 classes: \texttt{['Haghia Triada', 'Khania', 'Phaistos', 'Zakros', 'Knossos', 'Malia', 'Arkhanes']}.

\subsubsection{Machine Translation}

\subsection{Ancient Egyptian}

\subsubsection{Attribute classification}

We use time periods whose data more than 50 to conduct this task, resulting in 1,230 samples of 12 classes.

The classes are as below: \texttt{['Unas', 'Senwosret I Kheperkare', 'Pepi I Merire', 'Ramesses II Usermaatre-Setepenre', 'Sety I Menmaatre', 'Tuthmosis III Menkheperre (complete reign)', 'Pepi II Neferkare', 'Hatshepsut Maatkare', '18th Dynasty', 'Mentuhotep II Nebhepetre (complete reign)', 'Amenemhat II Nebukaure', 'Tutankhamun Nebkheperure', 'Cleopatra VII Philopator', '12th Dynasty']}

\subsection{Akkadian (Cuneiform)}

\subsubsection{Attribute prediction}

We use the dataset from \citet{chen2023cuneiml}, containing 34,562 photographs of Akkadian cuneiform tablets with their transcriptions in cuneiform unicode and Latin transliterations. The metadata contains attributes such as time period, genre and geographical location (found place) that we have used for attribute classification.

\subsubsection{Dependency parsing}

The data is from \citet{luukko-etal-2020-akkadian} and comes in as normalization form of Akkadian. We implemented a rule-based method to convert the normalization form back to standard transliteration for consistency. When the auto-conversion fails, we simply just keep the normalization form.

\subsubsection{Machine translation}

The data is from \citet{gutherz2023translating}, a collection of 8,056 lines of Akkadian-English pair. The Akkadian representation is available in both ATF transliteration and Cuneiform Unicode.

\subsection{Bamboo script}

There are actually more than one scripts used in bamboo slips, for example, Seal script and Clerical script. For simplicity, our dataset generally call it Bamboo script. We pick two famous collections of the bamboo script, BaoShan and GuoDian bamboo slip collections for attribute prediction and machine translation.

\subsubsection{Attribute prediction}

The BaoShan bamboo slips collection consists of 723 slips, which are in three genres.

\subsubsection{Machine Translation}

We used the GuoDian bamboo slips collection for machine translation, whose major genre is philosophical essay. By referring to \citep{liuzhaoguodian}, We manually labeled each complete sentences with their transliteration and translation in modern Chinese. After filtering out incomplete sentences or removing the sentences with high interpretation difficulty. We extracted and labelled 489 lines of parallel data. The Bamboo script was labeled by a trained interdisciplinary Ph.D. student under the guidance of Professor Wenbo Chen specializing in Ancient Chinese Philology.

\begin{figure}[t]
  \centering
  \includegraphics[width=0.95\linewidth]{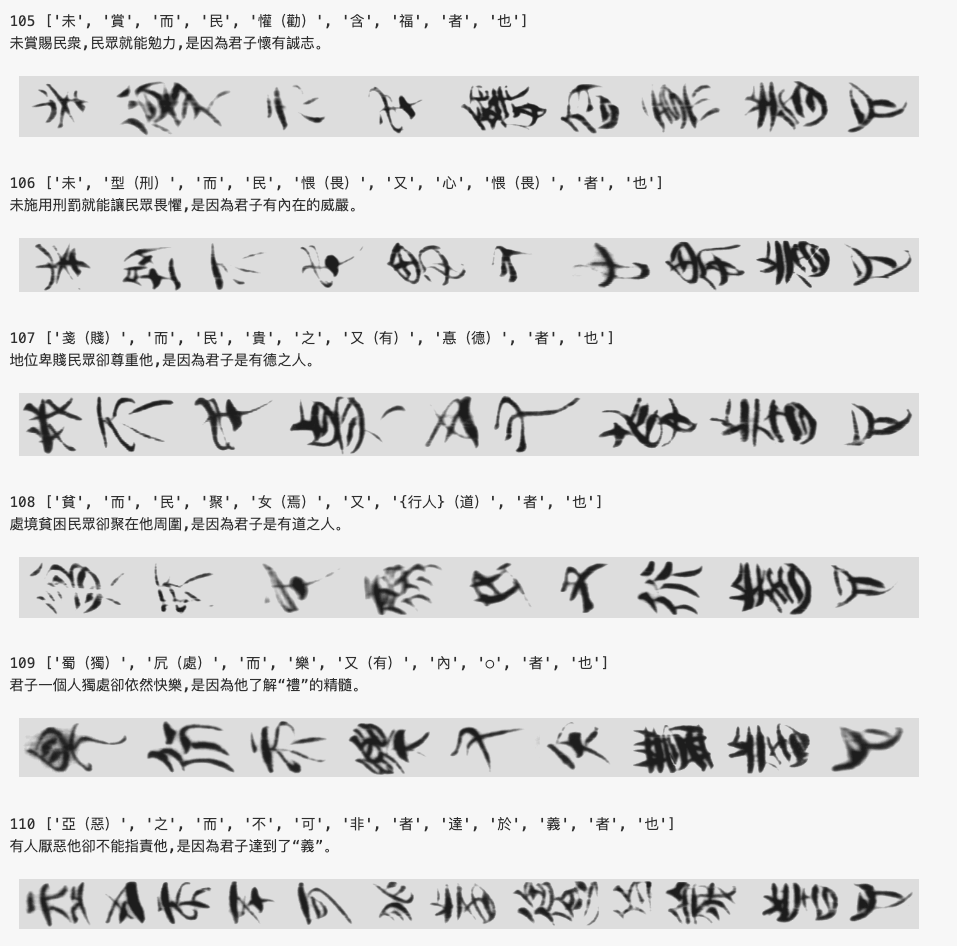}
  \caption{Sample parallel data of GuoDian Bamboo script dataset.}\label{fig:sample_bamboo}
\end{figure}




\section{Ablation Study on Text Recognition}\label{ch:detail_OCR}

\paragraph{Line-level OCR.}
We use Kraken \citep{Kiessling_The_Kraken_OCR_2022}, a state-of-the-art OCR library for historical documents, to transcribe the image into Latin or Unicode.
To handle unseen Unicode codepoints, we simply extend the vocabulary of the decoder.
For Latin transliteration, we predict outputs at the character level.
Similar to most OCR pipelines, the default OCR model of Kraken is a 2-layer bi-directional LSTM with a connectionist temporal classification (CTC) loss.

\paragraph{Glyph classification.}
We also applied Convolution Neural Networks (CNNs) to classify segmented glyphs.
We use ResNet-50 as the backbone model, followed by a linear classification layer.
We trained the model using cross-entropy loss with Adam optimizer.
We resize the image of each glyph to $64 \times 64$ and apply 20\% cropping.
In total, there are $21,687$ segmented characters in the ZHO dataset.

\section{More Translation Case Study}\label{sec:more_trans_examples}

We sample 10 examples from validation set for each language pairs.

\subsection{AKK - EN}
\begin{figure*}[h]
  \centering
  \includegraphics[width=\linewidth]{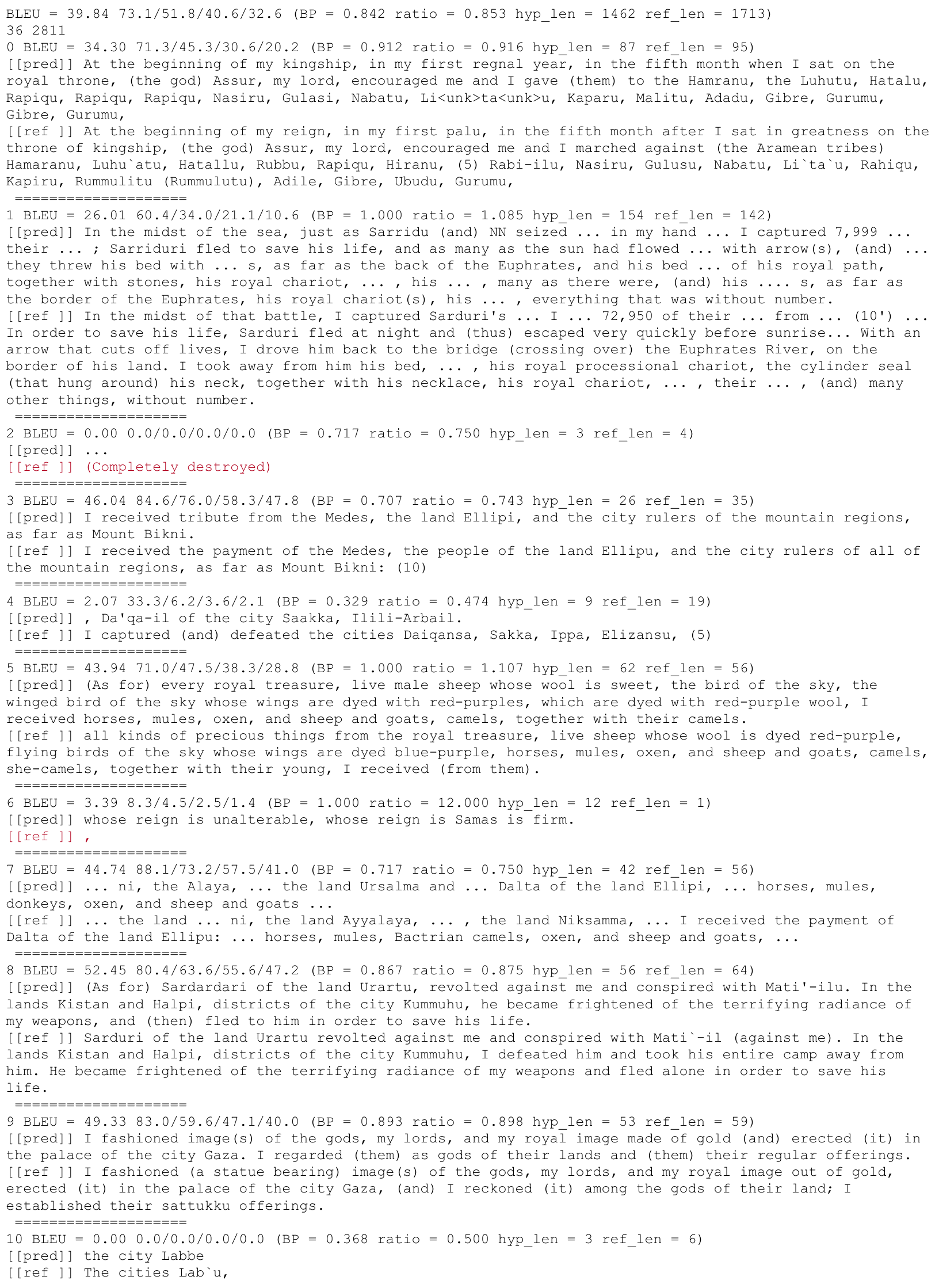}
  \caption{Akkadian translation samples from the PIXEL-MT models.}\label{fig:akk_trans}
\end{figure*}

Akkadian translation samples from the PIXEL-MT models is shown in \cref{fig:akk_trans}. We showcase the first 10 examples from the validation set, we find that some annotation error present for example \#2 and \#6.

\subsection{EGY - EN}

\begin{figure*}[h]
  \centering
  \includegraphics[width=\linewidth]{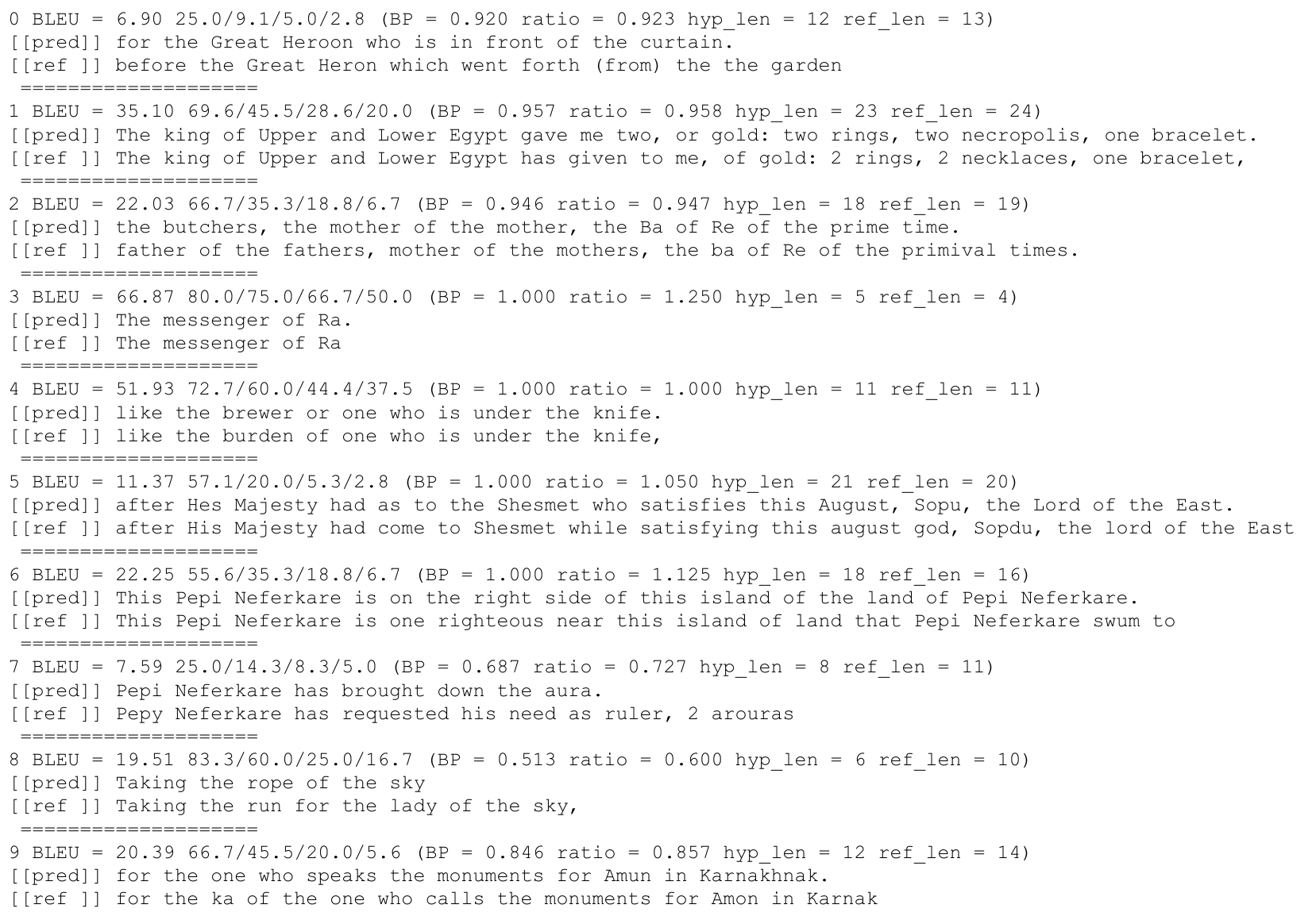}
  \caption{Ancient Egyptian translation samples from the PIXEL-MT models.}\label{fig:egy_trans}
\end{figure*}

Ancient Egyptian translation samples from the PIXEL-MT models is shown in \cref{fig:egy_trans}.

\subsection{ZHO - EN}

\begin{figure*}[h]
  \centering
  \includegraphics[width=\linewidth]{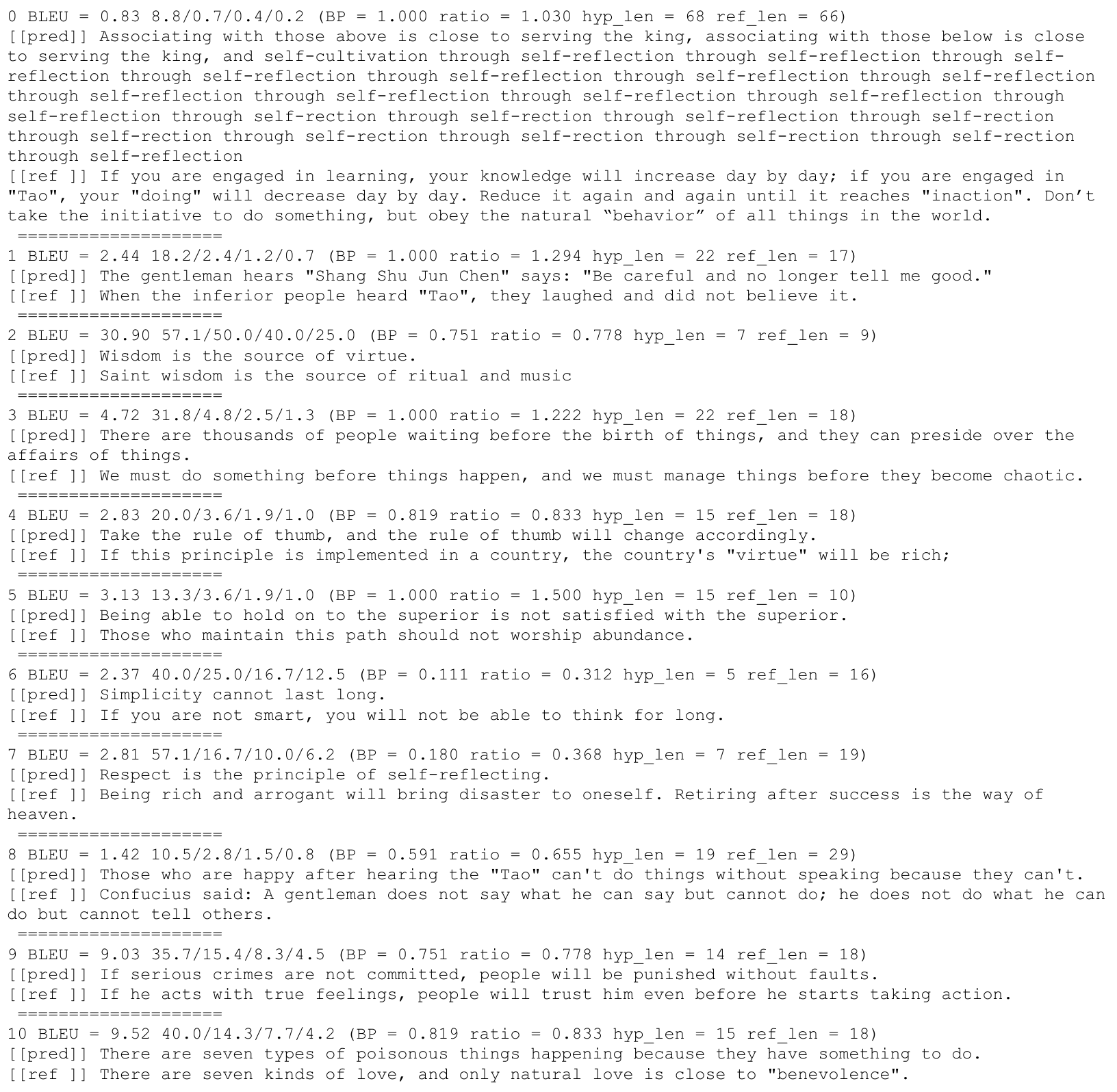}
  \caption{Old Chinese translation samples from the PIXEL-MT models.}\label{fig:zho_trans}
\end{figure*}

Old Chinese translation samples from the PIXEL-MT models is shown in \cref{fig:zho_trans}.

\end{document}